\documentclass[twoside]{article}

\usepackage[accepted]{aistats2022}
\usepackage[round]{natbib}

\usepackage[utf8]{inputenc} 
\usepackage[T1]{fontenc}    
\usepackage{url}            
\usepackage{booktabs}       
\usepackage{amsfonts}       
\usepackage{nicefrac}       
\usepackage{microtype}      
\usepackage{xcolor}         
\usepackage{makecell}
\usepackage{mathtools}
\usepackage[ruled]{algorithm}
\usepackage{algpseudocode}
\usepackage{subcaption}
\usepackage{wrapfig}
\usepackage{multirow}
\usepackage{amsmath}
\usepackage{enumitem}
\usepackage{IEEEtrantools}

\usepackage[pdftex]{graphicx}

\newcommand{\maximize}{\mathop{\rm maximize}}
\newcommand{\minimize}{\mathop{\rm minimize}}
\newcommand{\subjecto}{\mathop{\rm subject\;to}}

\newcommand{\argmax}{\mathop{\rm argmax}}

\newcommand{\bmu}{\ensuremath{{\boldsymbol \mu}}}

\newcommand{\bdelta}{\ensuremath{{\boldsymbol \delta}}}

\newcommand{\btheta}{\ensuremath{{\boldsymbol \theta}}}

\newtheorem{theorem}{Theorem}

\makeatletter
\newcommand{\algmargin}{\the\ALG@thistlm}
\makeatother
\newlength{\whilewidth}
\algdef{SE}[parWHILE]{parWhile}{EndparWhile}[1]
  {\parbox[t]{\dimexpr\linewidth-\algmargin}{%
     \hangindent\whilewidth\strut\algorithmicwhile\ #1\ \algorithmicdo\strut}}{\algorithmicend\ \algorithmicwhile}%
\algnewcommand{\parState}[1]{\State%
  \parbox[t]{\dimexpr\linewidth-\algmargin}{\strut #1\strut}}

\begin{document}

%

%

\twocolumn[

\aistatstitle{System-Agnostic Meta-Learning for MDP-based Dynamic Scheduling via Descriptive Policy}

\aistatsauthor{ Hyun-Suk Lee }

\aistatsaddress{ Sejong University } ]

\begin{abstract}
Dynamic scheduling is an important problem in applications from queuing to wireless networks.
It addresses how to choose an item among multiple scheduling items in each timestep to achieve a long-term goal.
Conventional approaches for dynamic scheduling find the optimal policy for a given specific system so that the policy from these approaches is usable only for the corresponding system characteristics.
Hence, it is hard to use such approaches for a practical system in which system characteristics dynamically change.
This paper proposes a novel policy structure for MDP-based dynamic scheduling, a descriptive policy, which has a system-agnostic capability to adapt to unseen system characteristics for an identical task (dynamic scheduling).
To this end, the descriptive policy learns a system-agnostic scheduling principle--in a nutshell, ``which condition of items should have a higher priority in scheduling''.
The scheduling principle can be applied to any system so that the descriptive policy learned in one system can be used for another system.
Experiments with simple explanatory and realistic application scenarios demonstrate that it enables system-agnostic meta-learning with very little performance degradation compared with the system-specific conventional policies.
\end{abstract}

\section{INTRODUCTION}
Dynamic scheduling is a general issue that has been important, from the past to the present, in various applications from recommender systems \citep{shani2005mdp,huang2021deep,lu2016partially} to communication networks \citep{ferra2003applying,wei2018user}.
Traditionally, dynamic scheduling problems have been often modeled based on Markov decision process (MDP) \citep{shani2005mdp,huang2021deep,lu2016partially,ferra2003applying,wei2018user,stidham1993survey,chang2000line,ye2018deep,xu2017deep,han2008resource} since scheduling is typically sequential decision-making.
To solve them, most conventional approaches find the optimal policy for a given system.
However, in practice, they may be useless since system characteristics dynamically change in general.

As a viable solution to this issue, meta-reinforcement learning (RL) can be considered  \citep{hospedales2020meta}.
Most state-of-the-art meta-RL methods focus on unseen tasks with identical dynamics \citep{lin2020model}.
On the other hand, meta-RL for dynamic scheduling should consider an identical task (scheduling) but unseen dynamics from the change of system characteristics.
To distinguish them, we name the meta-RL for dynamic scheduling \textit{system-agnostic} meta-learning since it addresses the different system dynamics.

\subsection{Motivation with MDP-based Dynamic Scheduling}
\label{sec:MDP}
We consider a general scheduling problem for dynamic systems consisting of a scheduler and multiple scheduling items (e.g., queue, users, devices, and so on) to be scheduled.
The scheduling items are indexed as $n\in\mathcal{N}=\{1,2,...,N\}$, where $N$ is the total number of the items in the system.
We assume a dynamic system that operates over a discrete time horizon $t\in\{1,2,...~\}$.
We call the information that describes the condition of the system feature information.
We define the vector of feature information of item $n$ in timestep $t$ as $f_n^t=(f_{n,1}^t,...,f_{n,K}^t)$, where $f_{n,k}^t$ denotes $k$-th feature information of item $n$ in timestep $t$ (e.g., current queue length in queue scheduling and channel condition in wireless networks) and $K$ is the number of types of features.
Then, the feature information in timestep $t$ represents the system state in timestep $t$.
Without loss of generality, we assume that each feature information is bounded by $[0,1]$.
In each timestep $t$, based on the system state, the scheduler determines an item to be served or to be processed and decisions relevant to scheduling (e.g., number of tasks to be served in job scheduling and transmission power in wireless networks) so as to achieve a goal of the system.
Then, the system state in the system varies according to the scheduler's decision and the random disturbances of the system.

Such a scheduling problem is a general one which has been widely used in various applications such as queueing \citep{ferra2003applying,stidham1993survey,chang2000line}, recommender systems \citep{shani2005mdp,huang2021deep,lu2016partially}, and state-of-the-art wireless networks \citep{wei2018user,ye2018deep,xu2017deep,han2008resource}.
To solve this problem, we typically formulate the problem as MDP.
First, a state in timestep $t$, $s^t\in\mathcal{S}$, is defined to explicitly provide the feature information of each item as\looseness=-1
\begin{equation}
\label{eqn:typical_state}
s^t=\left(f_1^t,f_2^t,...,f_N^t\right)\in\mathcal{S}=[0,1]^{N\times K},
\end{equation}
where $\mathcal{S}$ is a state space.
In addition, an action in timestep $t$, $a^t$, is defined to explicitly indicate the scheduling decisions as
\begin{equation}
\label{eqn:typical_action}
a^t=(n^t,m_1^t,...,m_L^t)\in\mathcal{A}=\mathcal{N}\times\mathcal{M}_1\times ...\times\mathcal{M}_L,
\end{equation}
where $n^t$ is the chosen item to be served or processed in timestep $t$ and $(m_1^t,...,m_L^t)$ are the decisions relevant to scheduling, and $L$ is the number of the relevant decisions.
Based on the state and action defined in the above, a policy $\pi:\mathcal{S}\rightarrow\mathcal{A}$ is defined to map states to actions.
To represent the goal of the system in a general form, we can use a utility function $u(s,a)$ which emits a reward signal relevant to the goal of the system according to the system state and the action.
The utility function deterministically calculates the utility according to the state and action, and the feature information of the items not chosen does not affect the utility.
Hence, if two items $n_1,n_2$ are in an identical condition (i.e., $f_{n_1}$ and $f_{n_2}$ are identical), whatever of them is chosen will have the same utility value.
Then, the system state transits according to the transition probabilities $\mathbb{P}(s^{t+1}|s^t,a^t)$, where the transition probabilities depend on the random disturbances in the system.
With the definitions, we can formulate an MDP-based dynamic scheduling problem as follows:
\begin{equation}
\label{eqn:mdp}
\maximize_{\pi:\mathcal{S}\rightarrow\mathcal{A}} ~U^\pi(s)=\mathbb{E}\left[\left. \sum_{t=0}^{\infty}{(\gamma)^t u(s^t,\pi(s^t))}\right|s^0=s \right],
\end{equation}
where $\gamma$ is a discount factor.
Then, we can define the optimal value function of the dynamic scheduling problem in \eqref{eqn:mdp} by
\begin{equation}
\label{eqn:mdp_optimal_value}
J^*(s)=\max_{\pi} U^\pi(s),~\forall s\in\mathcal{S}
\end{equation}
and its corresponding optimal policy by $\pi^*=\argmax_{\pi} U^\pi(s),~\forall s\in\mathcal{S}$.


In the literature, this problem formulation is typical and intuitive \citep{shani2005mdp,huang2021deep,lu2016partially,ferra2003applying,wei2018user,stidham1993survey,chang2000line,ye2018deep,xu2017deep,han2008resource} since in practice, most scheduling systems manage their corresponding items by using an index that can identify each item.
This problem can be solved by finding the optimal policy via standard dynamic programming \citep{bertsekas2000dynamic} or reinforcement learning approaches \citep{sutton2018reinforcement}.
However, the optimal policy is only for its specific given system with certain characteristics (e.g., the number of items and the statistical characteristics of the random disturbances).
Hence, if the system characteristics change, then the system becomes a new system at the point of the policy and the policy does not work anymore.
For example, we can find the following explicit reasons:
(1) The domain (i.e., the state space, $[0,1]^{N\times K}$) and codomain (i.e., the action space, $\mathcal{N}\times\mathcal{M}_1\times ...\times\mathcal{M}_L$) of the policy change according to the number of items, $N$;
(2) The optimal policy varies according to the statistical characteristics of the random disturbances.
In addition, the policy does not provide any information about its decision, which makes it harder to obtain the rationale of scheduling.
For convenience, we call such a policy in the conventional approaches a conventional policy in the rest of this paper.\looseness=-1

The dependency of the conventional policy on its corresponding system makes it harder to be used for a practical environment in which system characteristics dynamically change.
To address this issue, the concept of meta-learning can be adopted.
In an aspect of meta-learning, we need a meta-RL method for dynamic scheduling with different system characteristics, which can be described as a family of MDPs that share the \textit{identical task}--scheduling--but have \textit{different dynamics} from different system characteristics. We call this system-agnostic meta-learning.
However, typical meta-RL methods commonly address a family of MDPs that share the \textit{identical dynamics} but differ in the task specified by the reward function \citep{lin2020model}.
Hence, they are inappropriate to be used for dynamic scheduling whose goal is addressing unseen dynamics.
To clearly distinguish them, we can use a robot arm analogy in which the dynamics correspond to the movement of a robot arm with a given degree-of-freedom.
In this analogy, the typical meta-RL methods enable the robot arm to adapt to unseen tasks, such as picking an object and opening a drawer, by controlling the movement of the arm, but cannot address the robot arm with different degree-of-freedom (different dynamics).



\begin{figure*}[!tbp]
  \begin{subfigure}[b]{0.5\textwidth}
  \centering
    \includegraphics[width=0.9\textwidth]{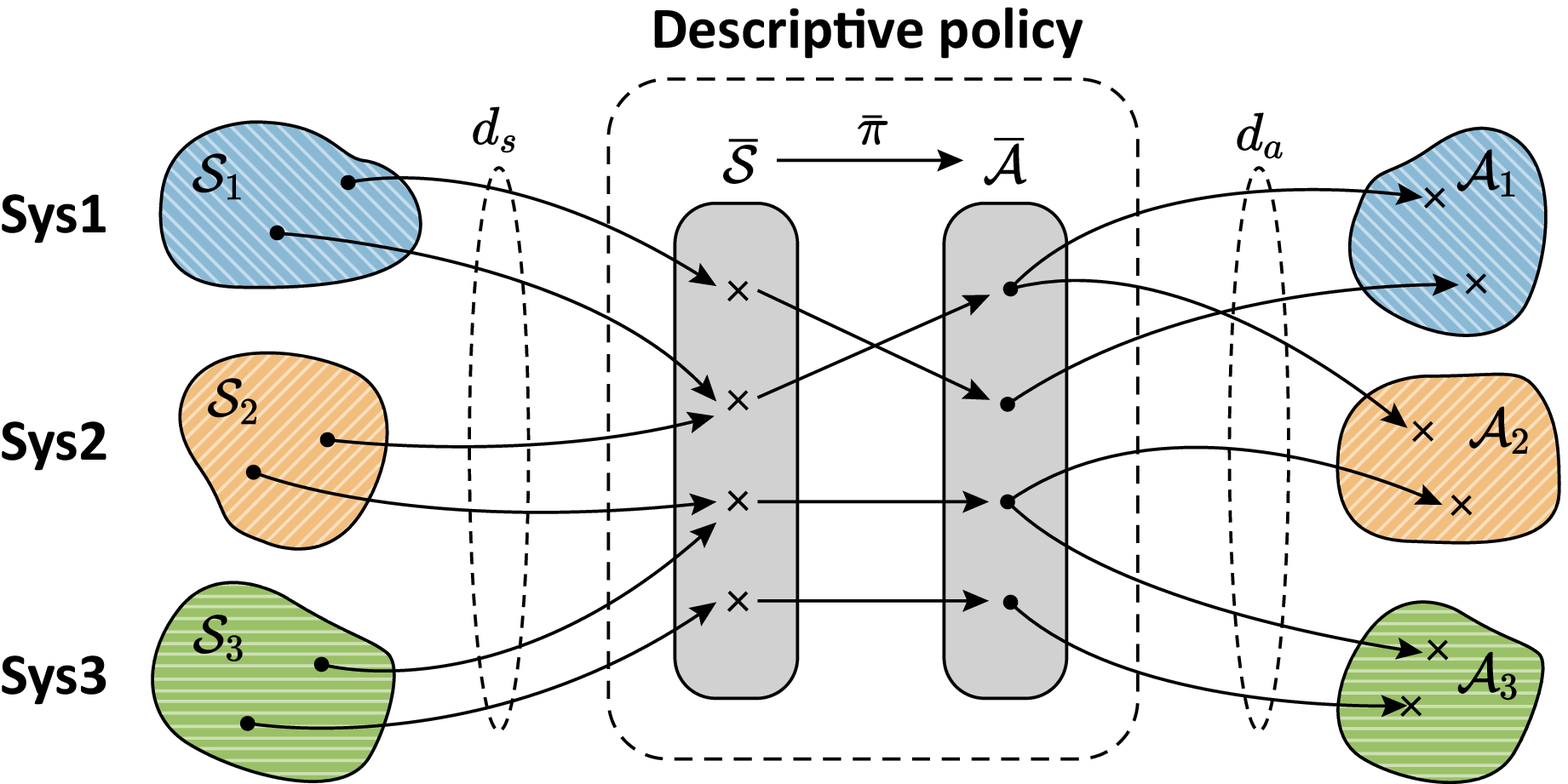}
    \caption{A system-agnostic descriptive policy.}
    \label{fig:descriptive_policy}
  \end{subfigure}
  \hfill
  \begin{subfigure}[b]{0.5\textwidth}
  \centering
    \includegraphics[width=0.85\textwidth]{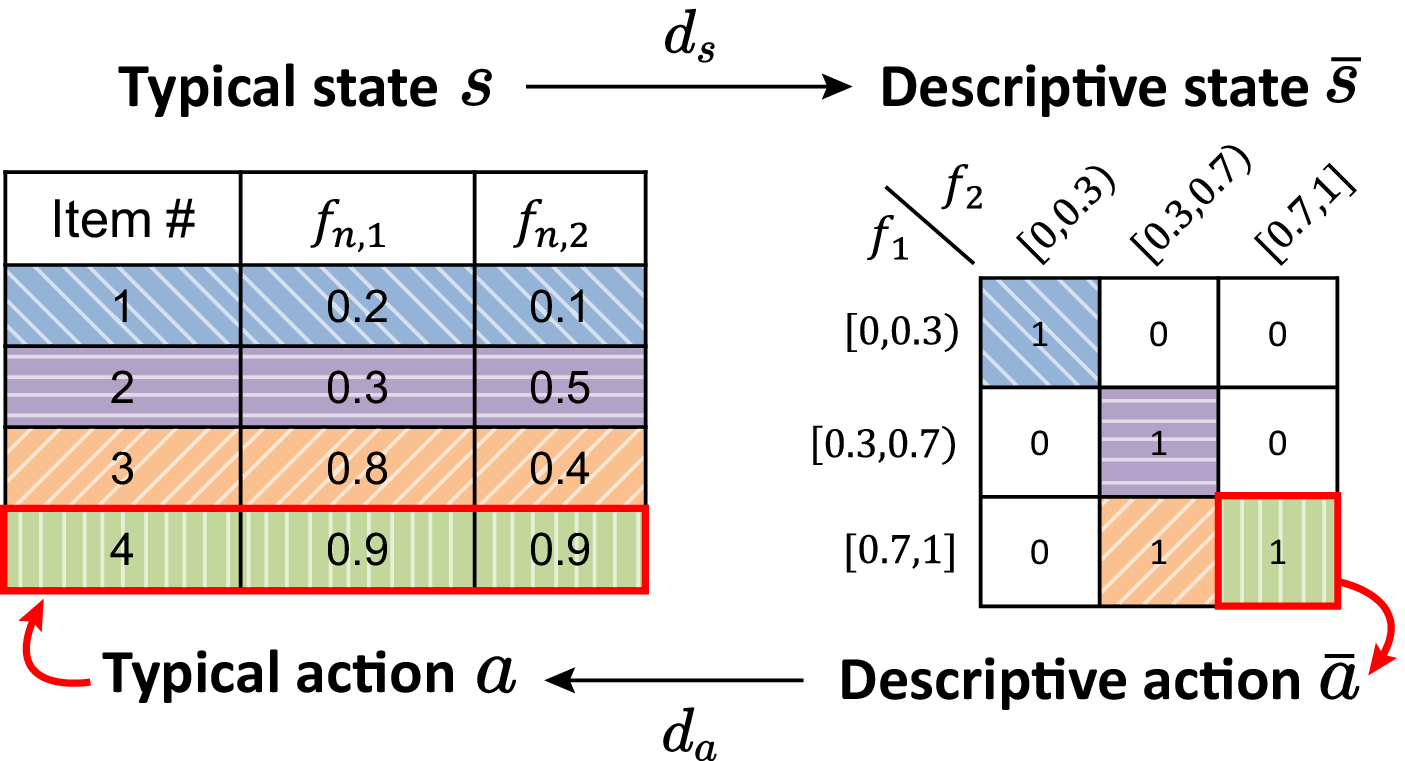}
    \caption{Descriptive state and action.}
    \label{fig:descriptive_state}
  \end{subfigure}
  \caption{Illustration of descriptive policy.\label{fig:illustration}}
\end{figure*}

\subsection{Our Contribution}
This paper proposes a novel descriptive policy for MDP-based dynamic scheduling to address the challenges of the conventional policy.
It has several distinctive characteristics illustrated in Fig. \ref{fig:illustration}.
\begin{itemize}[leftmargin=*]\itemsep=0pt
\item The definitions of states and actions for the descriptive policy focus on the condition of items rather than each item itself contrary to those for the conventional policy. This makes the descriptive policy system-agnostic as illustrated in Fig. \ref{fig:descriptive_policy}.

\item The descriptive policy learns a scheduling principle--``which condition of items should have a higher priority in scheduling''.
The scheduling principle can be applied to any system so that the descriptive policy learned in one system can be used for another system.

\item The descriptive policy provides the rationale of its decision thanks to its descriptive structure that describes the condition of the chosen item.
\end{itemize}
These characteristics of the descriptive policy enable system-agnostic meta-learning.
Experiments with simple explanatory and realistic application scenarios demonstrate that the descriptive policy is free from the burden of learning different policies for different systems
while achieving similar performance to the optimal policy.


\section{SYSTEM-AGNOSTIC META-LEARNING VIA DESCRIPTIVE POLICY}

\subsection{How to Design A Policy for System-Agnostic Meta-Learning?}
\label{sec:how_to}
The conventional policy structure based on the item index in \eqref{eqn:typical_state} and \eqref{eqn:typical_action} is intuitive, but at the same time, makes the conventional policy system-specific.
Hence, we need a novel policy structure that provides the system state and action without using the item index to design a system-agnostic policy.
Besides, to enable the system-agnostic policy, it should have the capability to learn a scheduling principle that can be applied to any system.
In a nutshell, the scheduling principle represents ``which \textit{condition} of items should have a higher priority in scheduling''.
With these backgrounds, we redesign the policy structure to focus on the \textit{condition} of each item rather than each \textit{item} itself in the conventional policy.
Specifically, in the novel policy structure, states describe the existence of users in a specific condition, and actions are specified by the conditions of users.
By this, it not only becomes system-agnostic but also learns the scheduling principle.

\subsection{Structure of Descriptive Policy}
\label{sec:descriptive_structure}
Here, we describe a structure of descriptive policy based on the MDP model in Section \ref{sec:MDP}.
\paragraph{Descriptive state}
For the descriptive policy, we first define a \textit{descriptive} state that represents \emph{whether an item that has a specific condition exists or not} of the feature information in each timestep.
To this end, we partition each $k$-th feature into multiple $H_k$ disjoint intervals and represent the condition as a combination of the intervals of each feature as illustrated in Fig. \ref{fig:descriptive_state}.
It is worth noting that discrete-valued features do not have to be partitioned and any partitioning methods can be used according to the characteristics of the feature.
We denote the index of the intervals in the partitions for $k$-th feature by $h_k\in\mathcal{H}_k=\{1,2,...,H_k\}$ and the interval $h_k$ by $D(h_k)$ (i.e., $\bigcup_{ h_k\in\mathcal{H}_k}D
(h_k)=[0,1]$).
We then define a \textit{descriptive} state in time-slot $t$, $\bar{s}^t$, by a $K$-dimensional matrix whose size is given by $\prod_{k\in\{1,...,K\}}H_k$.
Each element of the state whose index is given by a tuple $h=(h_1,...,h_K)$ has the value of 1 or 0 which indicates the existence of an item in the corresponding condition in the system
(i.e., the existence of an item satisfying $f_{n,k}^t\in D(h_k),~\forall k\in\{1,...,K\}$ for any $n$).
We denote it by
\begin{equation}
\label{eqn:desc_state}
\bar{s}^t(h)=\begin{cases} 1, &\mbox{if there exists any item in}\nonumber\\
&\mbox{the condition $h$ }\\
0, & \mbox{otherwise}
\end{cases}, 
\end{equation}
where $\bar{s}^t(h)$ represents the element of state $\bar{s}^t$ whose index is given by $h$.
Then, the state space is defined by $\bar{\mathcal{S}}=\{0,1\}^{\prod_{k\in\{1,...,K\}}H_k}$.
The descriptive state describes the system states in an implicit way focusing on the \textit{conditions} of the items.
This allows it to represent the system states in any system and take a look at the big picture of the system state to learn the scheduling principle.

\paragraph{Descriptive action}
In each timestep, the scheduler determines one scheduled item and the other scheduling decisions as in \eqref{eqn:typical_action}.
To avoid the dependency of actions to systems, we define a \textit{descriptive} action by using the descriptive state.
In specific, the descriptive action in timestep $t$, $\bar{a}^t$, is defined as the index of the element in the descriptive state and the other scheduling decisions,
\begin{equation}
\label{eqn:desc_action}
\bar{a}^t =(h_1^t,...,h_K^t,m_1^t,...,m_L^t)=(h^t,m^t).
\end{equation}
According to the descriptive action, the scheduler chooses an item in the condition corresponding to $h^t=(h_1^t,...,h_K^t)$ and serves the item with the scheduling decisions $m^t=(m_1^t,...,m_L^t)$.
Then, the action space is defined by $\bar{\mathcal{A}}=\prod_{ k\in\{1,...,K\}}\mathcal{H}_k\times \prod_{ l\in\{1,...,L\}}\mathcal{M}_l$.
In each timestep, the available actions depend on the descriptive state $\bar{s}^t$ since the action can be chosen only if an item exists in the condition of the chosen action.
Then, the set of feasible actions in timestep $t$ with state $\bar{s}^t$ is given by
\[
\bar{\mathcal{A}}(\bar{s}^t)=\{\bar{a}^t\in\bar{\mathcal{A}}| \bar{s}^t(h^t) = 1\}.
\]
The descriptive action can represent the scheduled item in any system by focusing on the condition of items rather than each item itself.
We illustrate the descriptive state and action in Fig. \ref{fig:descriptive_state}.


\paragraph{Descriptive Policy}
With the above ingredients, a descriptive policy can be defined as a mapping from the descriptive states to the descriptive actions, $\bar{\pi}:\bar{\mathcal{S}}\rightarrow\bar{\mathcal{A}}$.
This structure of the descriptive policy can be used in any system as illustrated in Fig. \ref{fig:descriptive_policy} since both descriptive state and action can represent the system states and scheduling decisions in any system.
Besides, its mapping from the descriptive states to actions describes which condition of items should be chosen in scheduling.
This implies that the descriptive policy has a capability to learn the scheduling principle.

\subsection{System-Agnostic Meta-Learning for MDP-based Dynamic Scheduling}
We propose a system-agnostic meta-learning approach in which the descriptive policy is learned via deep RL.
For simple presentation, we describe the proposed approach based on well-known deep Q-network (DQN) in \citep{mnih2015human}, but any other methods can be used as well.
To this end, we first consider a system that operates based on the definitions of typical states and actions in Section \ref{sec:MDP}.
We define the descriptive states and actions, $\bar{s}$ and $\bar{a}$, respectively, according to the definitions provided in Section \ref{sec:descriptive_structure}.
We then define a state translation function $d_s:\mathcal{S}\rightarrow\bar{\mathcal{S}}$ that translates a typical state $s^t$ in \eqref{eqn:typical_state} to a descriptive state $\bar{s}^t$ (i.e., $d_s(s^t)=\bar{s}^t$).
This can be done as in \eqref{eqn:desc_state}.
We also define an action translation function $d_a:\bar{\mathcal{A}}\rightarrow\mathcal{A}$ that translates a descriptive action $\bar{a}^t$ to a typical action $a^t$ (i.e., $d_a(\bar{a}^t)=a^t$).
We define a set of the indices of the items in condition $h$ as
$
\mathcal{N}(h)=\{n\in\mathcal{N}|f_{n,k}^t\in D(h_k),\forall k\in\{1,...,K\}\}.
$
Then, for a given $\bar{a}^t=(h^t,m^t)$, we have $d_a(\bar{a}^t)=(n^t,m^t)$, where $n^t$ is the index of the scheduled item that is arbitrarily chosen from $\mathcal{N}(h^t)$.
The role of the translation functions are illustrated in Fig. \ref{fig:descriptive_policy}.
We define the optimal value function based on the descriptive states and actions by
$
\bar{J}^*(\bar{s})=\max_{\bar{\pi}} \mathbb{E}\left[\left. \sum_{t=0}^\infty (\gamma)^t u(s^t,d_a(\bar{\pi}(\bar{s}^t))) \right|\bar{s}^0=\bar{s} \right],
$
where $\bar{\pi}$ is the descriptive policy and the utility function is the identical one in Section \ref{sec:MDP}.

\begin{algorithm}[H]
\caption{System-Agnostic Meta-Learning via Descriptive Policy}\label{alg:dqn}
\begin{algorithmic}[1]
\parState{Initialize DQN $\btheta$ and $t=1$}
\While{TRUE}
\State Choose $\bar{a}^t\in\bar{\mathcal{A}}(\bar{s}^t)$
\State Translate action $a^t\leftarrow d_a(\bar{a}^t)$
\State Observe $u^t$ and $s^{t+1}$
\State Translate state $\bar{s}^{t+1}\leftarrow d_s(s^{t+1})$
\State Store $(\bar{s}^t,\bar{a}^t,u^t,\bar{s}^{t+1})$
\State Update $\btheta$ using experiences
\State $t\leftarrow t+1$
\EndWhile
\end{algorithmic}
\end{algorithm}

To learn the optimal descriptive policy $\bar{\pi}^*$ via DQN, we implement a deep neural network composed of parameters $\btheta$ that approximate the optimal action-value function based on the descriptive states and actions, $\bar{Q}^*(\bar{s},\bar{a})$.\footnote{The action-value function based on the descriptive policy can be easily derived similar to the case of the value function.}
We denote the Q-approximation of given $\bar{s}$ and $\bar{a}$ using the DQN with $\btheta$ by $\bar{Q}(\bar{s},\bar{a};\btheta)$.
As in typical DQN, we can use the experiences from the system to train the DQN for the descriptive policy.
Algorithm \ref{alg:dqn} describes system-agnostic meta-learning via learning the descriptive policy.
For simple presentation, we consider only one episode here.
In timestep $t$, the descriptive action $\bar{a}^t$ is chosen from $\bar{\mathcal{A}}(\bar{s}^t)$ (line 3).
We can use any exploration method (e.g., a $\epsilon$-greedy method) to choose action.
Then, the chosen descriptive action should be translated into the typical one as $a^t=d_a(\bar{a}^t)$ (line 4).
If there exist multiple items having the condition specified by the descriptive action (i.e., $|\mathcal{N}(h^t)|>1$), one of them is arbitrarily selected.
According to action $a^t$, the system serves or processes the selected item.
Then, the system observes the utility, $u^t=u(s^t,a^t)$, and the state in timestep $t+1$, $s^{t+1}$ (line 5).
From the observed utility and the next state, an experience sample is stored in a form of a tuple $(\bar{s}^t,\bar{a}^t,u^t,\bar{s}^{t+1})$, where $\bar{s}^{t+1}=d_s(s^{t+1})$ (lines 6--7).
Then, the DQN can be trained to minimize its prediction error which is defined as
$
L(\btheta)=\left(u^t+\gamma\textstyle\max_{\bar{a}\in\bar{\mathcal{A}}(\bar{s}^{t+1})}Q(\bar{s}^{t+1},\bar{a};\btheta)-Q(\bar{s}^t,\bar{a}^t;\btheta)\right)^2
$ (line 8).
Typical techniques for effective learning such as experience replay and fixed-target Q-network can be adopted as in typical DQN.
By using this learning approach, we can train the descriptive policy in a system-agnostic way as in Fig. \ref{fig:descriptive_policy}.
As a result, it can be used for a practical system in which system characteristics dynamically changes, and it enables system-agnostic meta-learning.

\subsection{Theoretical Analysis}
\label{sec:theoretical_analysis}
For theoretical results, we first uniformly partition each feature into $2^b$ intervals to define a descriptive policy, where $b$ is the partitioning parameter.
Then, the length of each interval is given by $2^{-b}$.
We denote the optimal value function based on these intervals and the above definition of the descriptive state and action with the partitioning parameter $b$ by $\bar{J}^*_b(\bar{s})$.
We describe identical statistics of the random disturbances of items.
If two items $n_1,n_2$ have the identical statistics, then we have
$\mathbb{P}_{n_1}(f'|f,(k,m))=\mathbb{P}_{n_2}(f'|f,(k,m))$ and $\mathbb{P}_{n_1}(f'|f,(n_1,m))=\mathbb{P}_{n_2}(f'|f,(n_2,m)),~\forall f',f\in[0,1]^K,k\in\mathcal{N}\setminus\{n_1,n_2\},m\in\prod_{l\in\{1,...,L\}}\mathcal{M}_l$,
where $\mathbb{P}_k$ represents the probability over the random disturbance of item $k$ and the tuple denotes the action.
In the following theorem, the optimality of the descriptive policy is described.
\begin{theorem}
For a descriptive policy structure, if all items have identical statistics, then the optimal value of a given descriptive state $d_s(s)$ with parameter $b$, $\bar{J}^*_b(d_s(s))$, converges to the optimal value of a given typical state $s$, $J^*(s)$, as $b\rightarrow\infty$ (i.e., $J^*(s)=\lim_{b\rightarrow\infty} \bar{J}^*_b(d_s(s))$).
\end{theorem}
\begin{IEEEproof}
We here provide a sketch of the proof.
Suppose that $b$ is large enough that any two items in identical condition $h$ (i.e., belong to the same set $\mathcal{N}(h)$) are undistinguishable.
Then, from the assumption of the identical statistics, whichever choice the scheduler makes (choosing any item) via the action transformation function $d_a$, the system obtains identical results in a perspective of value function since the items have identical statistics and condition.
This implies that as $b\rightarrow\infty$, the optimal value function based on descriptive state and action converges to the optimal value function based on typical state and action.
The detailed proof is provided in the supplementary material.
\end{IEEEproof}


This theorem implies that the descriptive policy can achieve the optimal value if the statistics of all the items are identical and the partition is fine enough.
However, the statistics of each item are usually different from each other, and thus, for a given system, the descriptive policy is usually not optimal due to the generalization.
Nevertheless, the descriptive policy is still important since in real world, the system is not static.
This relationship will be discussed in the following section.

\subsection{Discussions on Descriptive Policy}

\begin{paragraph}{Relationship Between Descriptive Policy and System-Specific Policy}
In an aspect of machine learning, we can explain the relationship between the descriptive policy and the system-specific optimal policy as the well-known relation between a regularized model and an overfitted model \citep{goodfellow2016deep}.
For a given dataset used for training, the overfitted model typically achieves better performance than the regularized model. However, we need the regularized model that can be used in real world rather than the overfitted one that is better only for the given dataset.
This relation is similar to that between the descriptive policy and the system-specific optimal policy.
Typically, the system-specific optimal policy achieves better values to the given system than the descriptive policy, but it cannot be used for general use as in the overfitted model.
On the other hand, the descriptive policy can be used for any system by generalizing a policy for a given task--dynamic scheduling.
\end{paragraph}

\begin{paragraph}{Explainability of Descriptive Policy}
In XAI, the key point of explainability is providing a rationale of decision-making \citep{arrieta2020explainable,puiutta2020explainable}.
In the conventional policy, the index is the only information about the scheduled item that is provided from the action.
Thus, it is difficult to obtain any rationale for scheduling from the conventional policy.
On the other hand, the descriptive policy naturally provides the rationale of its decision thanks to its structure based on the scheduling principle.
In specific, the descriptive policy provides a rationale of why the scheduler chooses the scheduled item:
the condition of the scheduled item is more promising than the conditions of the other items to achieve a given goal.
This will be shown in experiments.
As with all other XAI technologies, such an explainability of the descriptive policy can be used to justify decision-making on behalf of interest groups.
However, the descriptive policy is relatively neutral compared with other XAI methods because it depends on reward only, and thus, it will become more important and critical.
\end{paragraph}

\section{EXTENSIONS OF DESCRIPTIVE POLICY}

\subsection{Dynamic Scheduling with Constraints}
\label{sec:extension_constraints}
We consider a dynamic scheduling problem with average constraints (e.g., average minimum utility per item and average minimum data rates of each user in wireless networks).
Such constraints transform the scheduling problem into a constrained MDP (CMDP) as follows:
\begin{equation}
\label{eqn:cmdp}
\arraycolsep=3pt\def\arraystretch{1.8}
\begin{array}{cl}\displaystyle
\maximize_{\pi:\mathcal{S}\rightarrow\mathcal{A}} & U^\pi(s) \\
\subjecto & U_n^\pi(s)\leq \delta_n,~\forall n\in\mathcal{N}
\end{array},
\end{equation}
where $U_n^\pi(s)=\mathbb{E}\left[\left. \sum_{t=0}^{\infty}{(\gamma)^t u_n(s^t,\pi(s^t))}\right|s^0=s \right]$, $u_n(s,a)$ is the utility of item $n$ related to its constraint, and $\delta_n$ is the constraint parameter of item $n$.
For simple presentation, here we consider only one constraint for each item, but it can be easily generalized to multiple constraints for each item.
To solve this CMDP, we reformulate it as the following unconstrained MDP (UMDP) by using the Lagrangian approach \citep{Altman1998Constrained}.
We first introduce a Lagrangian multiplier $\mu_n$ for the constraint of item $n$.
We then reformulate the CMDP to the UMDP as
\begin{equation}
\label{eqn:umdp}
\max_{\pi} \, \min_{\bmu\succeq\mathbf{0}} \, U^\pi(s)+\sum_{n\in\mathcal{N}}\mu_n(\delta_n-U_n^\pi(s)), \nonumber
\end{equation}
where $\bmu=(\mu_n)_{\forall n\in\mathcal{N}}$, $\succeq$ denotes the elementwise inequality operator, and $\mathbf{0}$ is the zero vector.
We can solve the CMDP by learning the optimal policy to the UMDP and the optimal Lagrangian multipliers $\bmu^*=(\mu^*_1,...,\mu^*_N)^\top$ together \citep{Mastronarde2013Joint}.
To find the optimal Lagrangian multipliers, we can use the stochastic subgradient algorithm \citep{ghadimi2012optimal} as
\begin{equation}\label{eqn:multiplier_update}
	\mu_n^{t+1}=\left[\mu_n^t-\alpha^t\left(\delta_n-u_n^t\right)\right]^+, ~ \forall n\in\mathcal{N},
\end{equation}
where $\alpha^t$ is the step size in time-slot $t$.
The detailed description of the Lagrangian approach to solve the scheduling problem with constraints is provided in the supplementary material.

To apply the descriptive policy structure for this CMDP, we now propose an augmented state $\underline{s}^t=(s^t,\bmu^t)$, where $\bmu^t=(\mu_1^t,...,\mu_N^t)^\top$.
The multiplier corresponding to item $n$ can be considered as a feature information describing the condition of item $n$.
According to the stochastic subgradient method in \eqref{eqn:multiplier_update}, the multiplier of the constraint of item $n$ decreases if the utility of item $n$, $u_n$, satisfies the constraint in each timestep and increases if not.
This implies that the multiplier of item $n$ represents the degree of unsatisfaction of its corresponding constraint.
Hence, by defining the descriptive state and action based on the augmented state, we can find a descriptive policy having a capability to satisfy the constraint of each item.
This will be shown via the experiments.


\subsection{Multiple Dynamic Scheduling Systems with Federated Learning}
Federated learning (FL) is one of the viable solutions for distributed learning \citep{mcmahan2017communication}.
It allows learning a shared global model at a global learner by aggregating the locally trained models of distributed local learners. 
Hence, under the coordination of the global learner, the local learners transmit their trained model to the global one instead of their local training data.
This enables training the global model in a distributed manner while avoiding privacy concerns because it does not require each local learner to upload its local training data.
In addition, FL has been widely used for deep RL as well to learn value networks and/or policy networks in a distributed way \citep{zhuo2019federated,wang2020federated}.

Here, we consider a scenario with multiple dynamic scheduling systems and a central server.
In this scenario, we can adopt FL as each system plays a role of a local learner and the central server aggregates the local policies from the systems as a global learner.
However, if the local policies follow the conventional policy structure, FL is difficult to be applied since each local policy considers different system characteristics.
Besides, the conventional policy structure makes each system have different neural network architectures for its local policy according to their systems, which implies that learning a global policy by simply aggregating the local policies is infeasible.
On the other hand, the descriptive policy structure naturally enables one to learn such a global policy via FL.
It not only makes the neural network architectures of each local policy identical but also enables each local system to learn the scheduling principle that can be used for any system.
Consequently, the descriptive policy structure allows us to enjoy FL for the multiple systems having identical scheduling tasks.
The experiment results for this are provided in the supplementary material.

%

\section{RELATED WORK}
\begin{paragraph}{Meta-RL}
Recently, meta-RL has been widely studied since meta-learning also known as ``learning to learn'' was established \citep{hospedales2020meta}.
Meta-RL tries to enable an agent to solve unseen tasks and environments efficiently.
To this end, meta-RL methods infer the task over a latent context \citep{lin2020model,rakelly2019efficient}, redesign policy neural network architecture \citep{duan2016rl,wang2016learning,LanLGW19}, control exploration strategies \citep{gupta2018meta}, and augment state or reward \citep{raileanu2020automatic,florensa2018automatic}.
These previous methods commonly focus on the family of tasks described by a family of MDPs which all share the same dynamics and environment but differ in the task specified by the reward function \citep{lin2020model}.
Hence, such meta-RL methods are not appropriate to address the target family of MDPs in this paper that share the same task but have different dynamics and environments. (Recall the robot arm analogy in Section \ref{sec:MDP}.)
Different from the methods above, our method can address this issue in dynamic scheduling by directly learning a \textit{principle} of the task that can be adapted to any dynamics.
To this end, it takes a novel descriptive policy structure that translates the different dynamics into a latent one.
\end{paragraph}

\begin{paragraph}{Explainable AI (XAI)}
With the great success of deep learning in recent, XAI has been widely studied to overcome a black-box nature of most deep learning models \citep{puiutta2020explainable}.
The black-box nature makes deep learning-based AI technologies including deep RL untrustworthy and harder to be used in practice due to a legal regulation \citep{arrieta2020explainable}.
Researches on XAI on deep RL are typically classified as transparent algorithms and post-hoc explainability \citep{arrieta2020explainable,heuillet2021explainability}.
The post-hoc explainability techniques have been widely studied to provide explanations of the trained policy via interaction with agent \citep{sequeira2020interestingness} and feature relevance analysis \citep{greydanus2018visualizing,Atrey2020Exploratory}.
On the other hand, transparent algorithms provide explanations while learning the policy.
To this end, the explanations are learned simultaneously with the policy in \citep{waa2018contrastive,madumal2020explainable}.
The methods in \citep{lesort2018state,zhang19m} learn representations of states, actions, or policy that are easier to handle than the original ones.
Despite their goal is not providing explainability, they can be useful for it \citep{heuillet2021explainability}.
The transparent algorithms via learning representations are the most related one to our method in the aspect of considering latent state/action structures.
However, different from them, our method is system-agnostic and does not require any additional burden for learning representations.
\end{paragraph}





\section{EXPERIMENTS}

In this section, we describe a variety of experiments using the descriptive policy in various scenarios.
We evaluate the descriptive policy by comparing its performance with the conventional policies that learn the system-specific optimal policies.
Moreover, we show its explainability as well.

\subsection{Simple Explanatory Scenario}
We first consider a simple explanatory scenario for our proposed system-agnostic meta-learning approach.
In the scenario, a dynamic scheduling system with $N$ items is considered.
We denote the price of item $n$ and the quantity of item $n$ to sell in timestep $t$ by $p_n^t\in[0,1]$ and $g_n^t\in\{0,1,2,3,4\}$, respectively.
In each timestep, $p_n$ and $g_n$ randomly vary in an i.i.d. manner and the system should select one of the items to sell it.
We assume that if an item is selected, it will be sold out during the scheduled timestep.
Then, the reward of timestep $t$ is derived as $r^t=p_{n^t}^tg_{n^t}^t$, where $n^t$ is the selected item in timestep $t$.
The goal of the scheduler is to maximize the reward.
In this scenario, the price and quantity of items to sell are the feature information used for the state.
The optimal policy for this scenario is trivial as a greedy policy to select the item maximizing the reward since there is no dependency between timesteps.
This implies that the optimal scheduling principle of this scenario is the following: choosing the item with larger $p_n$ and larger $g_n$ is more favorable to maximize the reward.

\begin{figure}[!tbp]\centering
  \begin{subfigure}[b]{0.48\textwidth}
    \includegraphics[width=\textwidth]{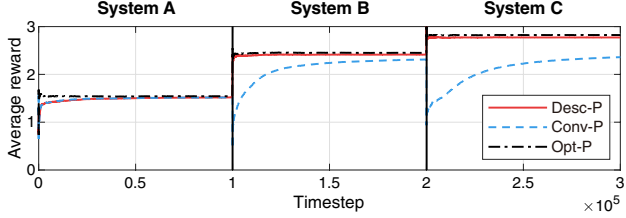}
    \caption{Average reward.}
    \label{fig:sc1_obj}
  \end{subfigure}
  \hfill
  \begin{subfigure}[b]{0.48\textwidth}
    \vspace{0.25in}
    \includegraphics[width=\textwidth]{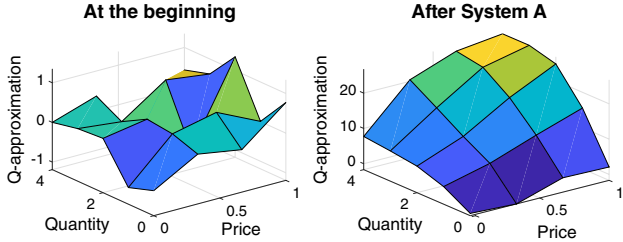}
    \caption{Q-approximations of the descriptive policy.}
    \label{fig:sc1_Q_value}
  \end{subfigure}
  \caption{Simulation results of simple explanatory scenario.\label{fig:sc1}}

\end{figure}

To show that the descriptive policy enables system-agnostic meta-learning, we consider three different system characteristics A, B, and C that have different numbers of items, $N$, as $\{2,6,10\}$, respectively.
For evaluation, we consider a conventional policy (Conv-P) for each system characteristics that is tailored to the system characteristics, a descriptive policy (Desc-P), and the optimal policy (Opt-P) for each system characteristics.
%
We consecutively perform the simulation for system characteristics A, B, and C in an online manner with $10^5$ time-slots for each system characteristics, but the descriptive policy is not trained during system characteristics C to clearly show its system-agnostic capability.
On the other hand, three Conv-Ps are trained during their corresponding system characteristics.
More details of the scenario including the simulation parameters are provided in the supplementary material.
The simulation results are provided in Fig. \ref{fig:sc1}.

In Fig. \ref{fig:sc1_obj}, the average rewards are provided.
To clearly show the effect of the change of the system characteristics, the reward is averaged for each system characteristics.
In the simulation, the system characteristics changes at timestep $10^5$ (from A to B) and $2\times 10^5$ (from B to C).
The Desc-P is not affected by the change, but the Conv-P should change the DQN at that time since the change of the system characteristics makes the policy structure different.
Hence, in the figure, we can see that the Desc-P effectively schedules achieving a close performance to the Opt-P even when the system characteristics change while the Conv-P undergoes an inefficiency in the early-stage of learning.
Moreover, the performance of the Desc-P in system characteristics C clearly shows that the Desc-P learns the scheduling principle that can be used in any system characteristics since the Desc-P schedules by using the experiences from system characteristics A and B only.

In Fig. \ref{fig:sc1_Q_value}, the Q-approximations of the Desc-P are provided.
The left plot in the figure illustrates the Q-approximation of the Desc-P at the beginning of the simulation while the right one illustrates that after learning with system characteristics A at timestep $10^5$.
At the beginning of the simulation, the Q-approximation does not have any trend.
On the other hand, after learning the experiences from system characteristics A, the Q-approximation has a clear trend that is aligned with the optimal scheduling principle.
As a result, such a trend makes the Desc-P achieve close performance to the optimal policy even in system characteristics C and clearly shows the system-agnostic capability of the Desc-P.
Besides, this trend explains the rationale of the decision from the policy as follows:
the item is chosen because it has larger $p_n$ and larger $g_n$ than the other items, which is more favorable to maximize the reward.
This shows the explainability of the Desc-P based on the scheduling principle.

\subsection{Realistic Application Scenario in Wireless Networks}
Here, we consider a realistic application scenario of user scheduling in a wireless network.
The network consists of one base station (BS) and multiple users that have minimum average data rate requirements.
The BS schedules a user and chooses a transmission power to serve the user over a discrete time horizon and in each timestep (typically, time-slot in wireless networks) the wireless channel condition of each user randomly varies as a random disturbance.
The goal of the scheduling is to minimize the average transmission power while satisfying the minimum average data rate requirements.
This scenario has a form of the dynamic scheduling problem with constraints described in Section \ref{sec:extension_constraints}.
In this system, the wireless channel condition and Lagrangian multiplier of the corresponding data rate requirement of each user are used as feature information.


In this scenario, we consider three different system characteristics that have different the number of users and their characteristics (the distance from the BS and the minimum average data rate requirements).
The simulation is conducted during $10^6$ time-slots for each system characteristics.
For system characteristics A and B (an online learning case), we consecutively perform the simulation as in the previous simple explanatory scenario.
For system characteristics C (a pre-trained case), we train a Conv-P in advance until it converges (about $10^7$ time-slots).
It is worth empasizing that the Conv-P is trained to learn the optimal policy tailored to system characteristics C.
Moreover, the Desc-P trained during system characteristics A and B is used as in the previous scenario to clearly show the system-agnostic capability.
More details of the scenario including the simulation parameters are provided in the supplementary material.

\begin{figure}[!tbp]
\centering
  \begin{subfigure}[b]{0.48\textwidth}
    \includegraphics[width=\textwidth]{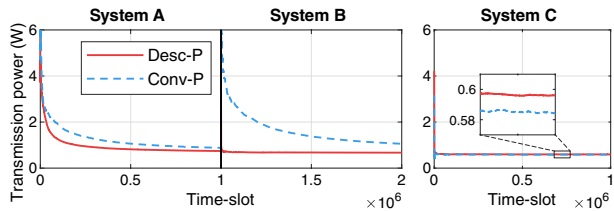}
    \caption{Average transmission power.}
    \label{fig:sc2_obj}
  \end{subfigure}
  \hfill
  \begin{subfigure}[b]{0.48\textwidth}
  \vspace{0.2in}
    \includegraphics[width=\textwidth]{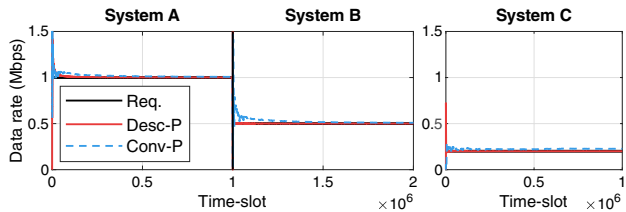}
    \caption{Average data rate of the user at a distance of 80 m.}
    \label{fig:sc2_rate}
  \end{subfigure}
  \caption{Simulation results of realistic application in wireless networks.\label{fig:sc2}}
\end{figure}

We provide the simulation results in the realistic scenario in wireless networks in Fig. \ref{fig:sc2}.
In Fig. \ref{fig:sc2_obj}, the average transmission power is provided.
The transmission power is averaged for each system characteristics as in the previous scenario.
In the online learning case, the system characteristics change from A to B at time-slot $10^6$.
As in the previous results, we can see that the Desc-P is not affected by that change, but the Conv-P should be changed at that time and goes through the early-stage of learning.
In the pre-trained case, we can see that the Desc-P achieves a close performance with the Conv-P.
This clearly shows that the Desc-P still learns the scheduling principle in the realistic application scenario.

Fig. \ref{fig:sc2_rate} provides the average data rate of the user at a distance of 80 m in each system characteristics since the user is the farthest one from the BS, which implies the most challenging user to satisfy the data rate requirement among the users.
The result shows that the Desc-P satisfies the data rate requirements in all system characteristics.
It is worth emphasizing that system characteristics C has different requirements from those in system characteristics A and B and the Desc-P is trained by only using the experiences from system characteristics A and B.
This clearly shows that the extended descriptive policy described in Section \ref{sec:extension_constraints} learns the scheduling principle to satisfy the requirements.

\section{CONCLUSION}
In this paper, we have studied explainable system-agnostic meta-learning for MDP-based dynamic scheduling. The policy proposed, descriptive policy, describes the condition of items in the state and action, which makes the policy system-agnostic and explainable. The descriptive policy enables system-agnostic meta-learning for dynamic scheduling by learning the scheduling principle that can be used in any system. Experiments with simple explanatory and realistic application scenarios demonstrate that the descriptive policy achieves similar performance to the optimal one while enjoying its advantages.

\subsubsection*{Acknowledgements}
This work was supported by the National Research Foundation of Korea (NRF) grant through the Korea Government (MSIT) under Grant 2021R1G1A1004796. We thank all reviewers for their comments and suggestions.

\bibliography{mybib}

\begin{thebibliography}{}

\bibitem[Altman, 1998]{Altman1998Constrained}
Altman, E. (1998).
\newblock {\em Constrained Markov Decision Processes}.
\newblock Chapman and Hall/CRC.

\bibitem[Arrieta et~al., 2020]{arrieta2020explainable}
Arrieta, A.~B., D{\'\i}az-Rodr{\'\i}guez, N., Del~Ser, J., Bennetot, A., Tabik,
  S., Barbado, A., Garc{\'\i}a, S., Gil-L{\'o}pez, S., Molina, D., Benjamins,
  R., et~al. (2020).
\newblock Explainable artificial intelligence ({XAI}): Concepts, taxonomies,
  opportunities and challenges toward responsible {AI}.
\newblock {\em Information Fusion}, 58:82--115.

\bibitem[Atrey et~al., 2020]{Atrey2020Exploratory}
Atrey, A., Clary, K., and Jensen, D. (2020).
\newblock Exploratory not explanatory: Counterfactual analysis of saliency maps
  for deep reinforcement learning.
\newblock In {\em Proceedings of the International Conference on Learning
  Representations (ICLR)}.

\bibitem[Bertsekas et~al., 2000]{bertsekas2000dynamic}
Bertsekas, D.~P. et~al. (2000).
\newblock {\em Dynamic programming and optimal control: Vol. 1}.
\newblock Athena scientific Belmont.

\bibitem[Chang et~al., 2000]{chang2000line}
Chang, H.~S., Givan, R., and Chong, E.~K. (2000).
\newblock On-line scheduling via sampling.
\newblock In {\em Proceedings of the International Conference on Artificial
  Intelligence Planning and Scheduling (AIPS)}, pages 62--71.

\bibitem[Duan et~al., 2016]{duan2016rl}
Duan, Y., Schulman, J., Chen, X., Bartlett, P.~L., Sutskever, I., and Abbeel,
  P. (2016).
\newblock {RL}$^2$: Fast reinforcement learning via slow reinforcement
  learning.
\newblock {\em arXiv preprint arXiv:1611.02779}.

\bibitem[Ferr{\'a} et~al., 2003]{ferra2003applying}
Ferr{\'a}, H.~L., Lau, K., Leckie, C., and Tang, A. (2003).
\newblock Applying reinforcement learning to packet scheduling in routers.
\newblock In {\em Proceedings of The Innovative Applications Conference on
  Artificial Intelligence (IAAI)}, pages 79--84.

\bibitem[Florensa et~al., 2018]{florensa2018automatic}
Florensa, C., Held, D., Geng, X., and Abbeel, P. (2018).
\newblock Automatic goal generation for reinforcement learning agents.
\newblock In {\em Proceedings of the International Conference on Machine
  Learning (ICML)}, pages 1515--1528. PMLR.

\bibitem[Ghadimi and Lan, 2012]{ghadimi2012optimal}
Ghadimi, S. and Lan, G. (2012).
\newblock Optimal stochastic approximation algorithms for strongly convex
  stochastic composite optimization {I}: A generic algorithmic framework.
\newblock {\em SIAM Journal on Optimization}, 22(4):1469--1492.

\bibitem[Goodfellow et~al., 2016]{goodfellow2016deep}
Goodfellow, I., Bengio, Y., and Courville, A. (2016).
\newblock {\em Deep learning}.
\newblock MIT press.

\bibitem[Greydanus et~al., 2018]{greydanus2018visualizing}
Greydanus, S., Koul, A., Dodge, J., and Fern, A. (2018).
\newblock Visualizing and understanding atari agents.
\newblock In {\em Proceedings of the International Conference on Machine
  Learning (ICML)}, pages 1792--1801. PMLR.

\bibitem[Gupta et~al., 2018]{gupta2018meta}
Gupta, A., Mendonca, R., Liu, Y., Abbeel, P., and Levine, S. (2018).
\newblock Meta-reinforcement learning of structured exploration strategies.
\newblock In {\em Advances in Neural Information Processing Systems (NeurIPS)},
  pages 5307--5316.

\bibitem[Han and Liu, 2008]{han2008resource}
Han, Z. and Liu, K.~R. (2008).
\newblock {\em Resource allocation for wireless networks: basics, techniques,
  and applications}.
\newblock Cambridge university press.

\bibitem[Heuillet et~al., 2021]{heuillet2021explainability}
Heuillet, A., Couthouis, F., and D{\'\i}az-Rodr{\'\i}guez, N. (2021).
\newblock Explainability in deep reinforcement learning.
\newblock {\em Knowledge-Based Systems}, 214:106685.

\bibitem[Hospedales et~al., 2020]{hospedales2020meta}
Hospedales, T., Antoniou, A., Micaelli, P., and Storkey, A. (2020).
\newblock Meta-learning in neural networks: A survey.
\newblock {\em arXiv preprint arXiv:2004.05439}.

\bibitem[Huang et~al., 2021]{huang2021deep}
Huang, L., Fu, M., Li, F., Qu, H., Liu, Y., and Chen, W. (2021).
\newblock A deep reinforcement learning based long-term recommender system.
\newblock {\em Knowledge-Based Systems}, 213:106706.

\bibitem[Lan et~al., 2019]{LanLGW19}
Lan, L., Li, Z., Guan, X., and Wang, P. (2019).
\newblock Meta reinforcement learning with task embedding and shared policy.
\newblock In {\em Proceedings of the International Joint Conference on
  Artificial Intelligence (IJCAI)}, pages 2794--2800.

\bibitem[Lesort et~al., 2018]{lesort2018state}
Lesort, T., D{\'\i}az-Rodr{\'\i}guez, N., Goudou, J.-F., and Filliat, D.
  (2018).
\newblock State representation learning for control: An overview.
\newblock {\em Neural Networks}, 108:379--392.

\bibitem[Lin et~al., 2020]{lin2020model}
Lin, Z., Thomas, G., Yang, G., and Ma, T. (2020).
\newblock Model-based adversarial meta-reinforcement learning.
\newblock {\em Advances in Neural Information Processing Systems (NeurIPS)},
  33.

\bibitem[Lu and Yang, 2016]{lu2016partially}
Lu, Z. and Yang, Q. (2016).
\newblock Partially observable markov decision process for recommender systems.
\newblock {\em arXiv preprint arXiv:1608.07793}.

\bibitem[Madumal et~al., 2020]{madumal2020explainable}
Madumal, P., Miller, T., Sonenberg, L., and Vetere, F. (2020).
\newblock Explainable reinforcement learning through a causal lens.
\newblock In {\em Proceedings of the AAAI Conference on Artificial Intelligence
  (AAAI)}, volume~34, pages 2493--2500.

\bibitem[Mastronarde and van~der Schaar, 2013]{Mastronarde2013Joint}
Mastronarde, N. and van~der Schaar, M. (2013).
\newblock Joint physical-layer and system-level power management for
  delay-sensitive wireless communications.
\newblock {\em IEEE Transactions on Mobile Computing}, 12(4):694--709.

\bibitem[McMahan et~al., 2017]{mcmahan2017communication}
McMahan, B. et~al. (2017).
\newblock Communication-efficient learning of deep networks from decentralized
  data.
\newblock In {\em Proceedings of the International Conference on Artificial
  Intelligence and Statistics (AISTATS)}.

\bibitem[Mnih et~al., 2015]{mnih2015human}
Mnih, V., Kavukcuoglu, K., Silver, D., Rusu, A.~A., et~al. (2015).
\newblock Human-level control through deep reinforcement learning.
\newblock {\em Nature}, 518(7540):529--533.

\bibitem[Puiutta and Veith, 2020]{puiutta2020explainable}
Puiutta, E. and Veith, E.~M. (2020).
\newblock Explainable reinforcement learning: A survey.
\newblock In {\em Proceedings of the International Cross-Domain Conference for
  Machine Learning and Knowledge Extraction (CD-MAKE)}, pages 77--95. Springer.

\bibitem[Raileanu et~al., 2020]{raileanu2020automatic}
Raileanu, R., Goldstein, M., Yarats, D., Kostrikov, I., and Fergus, R. (2020).
\newblock Automatic data augmentation for generalization in deep reinforcement
  learning.
\newblock {\em arXiv preprint arXiv:2006.12862}.

\bibitem[Rakelly et~al., 2019]{rakelly2019efficient}
Rakelly, K., Zhou, A., Finn, C., Levine, S., and Quillen, D. (2019).
\newblock Efficient off-policy meta-reinforcement learning via probabilistic
  context variables.
\newblock In {\em Proceedings of the International Conference on Machine
  Learning (ICML)}, pages 5331--5340. PMLR.

\bibitem[Sequeira and Gervasio, 2020]{sequeira2020interestingness}
Sequeira, P. and Gervasio, M. (2020).
\newblock Interestingness elements for explainable reinforcement learning:
  Understanding agents' capabilities and limitations.
\newblock {\em Artificial Intelligence}, 288:103367.

\bibitem[Shani et~al., 2005]{shani2005mdp}
Shani, G., Heckerman, D., Brafman, R.~I., and Boutilier, C. (2005).
\newblock An {MDP}-based recommender system.
\newblock {\em Journal of Machine Learning Research}, 6(9).

\bibitem[Stidham and Weber, 1993]{stidham1993survey}
Stidham, S. and Weber, R. (1993).
\newblock A survey of markov decision models for control of networks of queues.
\newblock {\em Queueing systems}, 13(1):291--314.

\bibitem[Sutton and Barto, 2018]{sutton2018reinforcement}
Sutton, R.~S. and Barto, A.~G. (2018).
\newblock {\em Reinforcement learning: An introduction}.
\newblock MIT press.

\bibitem[Waa et~al., 2018]{waa2018contrastive}
Waa, J., Diggelen, J.~v., Bosch, K., and Neerincx, M. (2018).
\newblock Contrastive explanations for reinforcement learning in terms of
  expected consequences.
\newblock In {\em Proceedings of the Workshop on Explainable AI on the IJCAI
  conference, Stockholm, Sweden., 37}.

\bibitem[Wang et~al., 2016]{wang2016learning}
Wang, J.~X., Kurth-Nelson, Z., Tirumala, D., Soyer, H., Leibo, J.~Z., Munos,
  R., Blundell, C., Kumaran, D., and Botvinick, M. (2016).
\newblock Learning to reinforcement learn.
\newblock {\em arXiv preprint arXiv:1611.05763}.

\bibitem[Wang et~al., 2020]{wang2020federated}
Wang, X., Wang, C., Li, X., Leung, V.~C., and Taleb, T. (2020).
\newblock Federated deep reinforcement learning for internet of things with
  decentralized cooperative edge caching.
\newblock {\em IEEE Internet of Things Journal}, 7(10):9441--9455.

\bibitem[Wei et~al., 2018]{wei2018user}
Wei, Y., Yu, F.~R., Song, M., and Han, Z. (2018).
\newblock User scheduling and resource allocation in {HetNets} with hybrid
  energy supply: An actor-critic reinforcement learning approach.
\newblock {\em IEEE Transactions on Wireless Communications}, 17(1):680--692.

\bibitem[Xu et~al., 2017]{xu2017deep}
Xu, Z., Wang, Y., Tang, J., Wang, J., and Gursoy, M.~C. (2017).
\newblock A deep reinforcement learning based framework for power-efficient
  resource allocation in cloud {RANs}.
\newblock In {\em Proceedings of the IEEE International Conference on
  Communications (ICC)}.

\bibitem[Ye et~al., 2019]{ye2018deep}
Ye, H., Li, G.~Y., and Juang, B.-H.~F. (2019).
\newblock Deep reinforcement learning based resource allocation for v2v
  communications.
\newblock {\em IEEE Transactions on Vehicular Technology}, 68(4):3163--3173.

\bibitem[Zhang et~al., 2019]{zhang19m}
Zhang, M., Vikram, S., Smith, L., Abbeel, P., Johnson, M., and Levine, S.
  (2019).
\newblock {SOLAR}: Deep structured representations for model-based
  reinforcement learning.
\newblock In {\em Proceedings of the International Conference on Machine
  Learning (ICML)}, pages 7444--7453.

\bibitem[Zhuo et~al., 2019]{zhuo2019federated}
Zhuo, H.~H., Feng, W., Xu, Q., Yang, Q., and Lin, Y. (2019).
\newblock Federated deep reinforcement learning.
\newblock {\em arXiv preprint arXiv:1901.08277}.

\end{thebibliography}
\bibliographystyle{unsrt}


\clearpage
\appendix

\thispagestyle{empty}

\onecolumn \makesupplementtitle

\section{PROOF OF THEOREM 1}

To prove Theorem 1, we first define an occupancy measure for policy $\pi$ with initial state $s$ and state-action pair $(s',a')$
\[
f^\pi(s;s',a')=(1-\gamma)\sum_{t=1}^\infty \gamma^{t-1}\phi^\pi(s^t=s',a^t=a),
\]
where $\phi^\pi(s,a)$ is a probability measure over the state and action trajectories for policy $\pi$.
The occupancy measure denotes the proportion of time that the MDP visits state-action pair $(s',a')$ in the long run according to the policy $\pi$ with initial state $s$.
By using this, the value function of policy $\pi$ can be written as a linear combination of the occupancy measure and the utility for all the possible state-action pairs as follows:
\begin{equation}
\label{eqn:pf1_0}
J^\pi(s)=\sum_{(s',a')}f^\pi(s;s',a')u(s',a').
\end{equation}
Then, the optimal value function can be written as
$
J^*(s)=\sum_{(s',a')}f^{\pi^*}(s;s',a')u(s',a'),
$
where $\pi^*$ is the optimal policy.

Let $s$ be a state of the dynamic scheduling system in which when translating it into descriptive state $\bar{s}$, at least one condition $h$ having two or more items exists.
We denote any two items in identical condition $h$ (i.e., belong to the same set $\mathcal{N}(h)$) by $n_1$ and $n_2$.
The difference between the utility with scheduling $n_1$ and that with scheduling $n_2$ is defined as
\begin{equation}
\label{eqn:pf1_1}
D(n_1,n_2)=|u(s,n_1,m)-u(s,n_2,m)|=|u(f_{n_1},n_1,m)-u(f_{n_2},n_2,m)|,
\end{equation}
where $(n,m)$ denotes the action and $f_n$ is the feature information of item $n$.
The second equality in \eqref{eqn:pf1_1} holds since the utility is determined by only the feature information of the scheduled item.
Then, $D(n_1,n_2)$ converges to zero as $b$ goes to infinity (i.e., $f_{n_1}=f_{n_2}$ and	 $u(s,n_1,m)=u(s,n_2,m)$ when $b=\infty$) since the utility is deterministic.
%
%
%
Also, we consider the state $s'$ subsequent to the state $s$.
With $b=\infty$, for any $s'=(f'_{1},...,f'_{n_1},...,f'_{n_2},...,f'_{N})$, we have 
\[
\mathbb{P}(f'_{1},...,f'_{n_1},...,f'_{n_2},...,f'_{N}|s,(n_1,m))=
\mathbb{P}(f'_{1},...,f'_{n_2},...,f'_{n_1},...,f'_{N}|s,(n_2,m)) \textrm{ and }
\]
\[
\mathbb{P}(f'_{1},...,f'_{n_2},...,f'_{n_1},...,f'_{N}|s,(n_1,m))=
\mathbb{P}(f'_{1},...,f'_{n_1},...,f'_{n_2},...,f'_{N}|s,(n_2,m)),
\]
since we assume the identical statistics of the random disturbances of the items (i.e., $\mathbb{P}_{n_1}(f'|f,(k,m))=\mathbb{P}_{n_2}(f'|f,(k,m))$ and $\mathbb{P}_{n_1}(f'|f,(n_1,m))=\mathbb{P}_{n_2}(f'|f,(n_2,m)),~\forall f',f\in[0,1]^K,k\in\mathcal{N}\setminus\{n_1,n_2\},m\in\prod_{l\in\{1,...,L\}}\mathcal{M}_l$, where $\mathbb{P}_k$ represents the probability over the random disturbance of item $k$.
This implies that the state-action trajectories according to a given policy after scheduling $n_1$ or $n_2$ are undistinguishable.

These ingredients imply that scheduling the items in the same condition are undistinguishable when $b=\infty$ in a perspective of value function which is the linear combination of the utilities and the occupancy measures in \eqref{eqn:pf1_0}.
Then, we can find a descriptive policy that is undistinguishable from the optimal policy $\pi^*$ in a perspective of value function.
For given state $s$, if the optimal action $\pi^*(s)$ does not have any other items with the identical condition, then the descriptive policy can uniquely represent it.
If not, then the descriptive policy can represent the action that is undistinguishable to $\pi^*(s)$ in a perspective of value function.
Consequently, the descriptive policy that represent the optimal policy $\pi^*$ is the optimal descriptive policy $\bar{\pi}^*$ when $b=\infty$, and its value function is identical to the optimal one, $J^*(s)$.
Hence, the value function of the optimal descriptive policy with parameter $b$, $\bar{J}^*_b$, becomes identical to the optimal value function, $J^*(s)$, as $b\rightarrow\infty$.


\section{SOLUTION TO DYNAMIC SCHEDULING PROBLEM WITH CONSTRAINTS}
We consider a dynamic scheduling problem with average constraints (e.g., average minimum utility per item and average minimum data rates of each user in wireless networks).
Such constraints transform the scheduling problem into a constrained MDP (CMDP) as follows:
\begin{equation}
\label{eqn:cmdp}
\maximize_{\pi:\mathcal{S}\rightarrow\mathcal{A}} ~U^\pi(s)~
\subjecto ~U_n^\pi(s)\leq \delta_n,~\forall n\in\mathcal{N},
\end{equation}
where $U_n^\pi(s)=\mathbb{E}\left[\left. \sum_{t=0}^{\infty}{(\gamma)^t u_n(s^t,\pi(s^t))}\right|s^0=s \right]$, $u_n(s,a)$ is the utility of item $n$ related to its constraint, and $\delta_n$ is the constraint parameter of item $n$.
For simple presentation, here we consider only one constraint for each item, but it can be easily generalized to multiple constraints for each item.
The optimal value of the CMDP in \eqref{eqn:cmdp} is defined as
\[
J^*(s)=\max_{\pi:U_n^\pi\leq\delta_n,\forall n\in\mathcal{N}}U^\pi(s),~\forall	s\in\mathcal{S}.
\]
To solve this CMDP, we reformulate it as the following unconstrained MDP (UMDP) by using the Lagrangian approach \citep{Altman1998Constrained}.
We first introduce a Lagrangian multiplier $\mu_n$ for the constraint of item $n$.
We then define a Lagrangian utility function for action $a$ and state $s$ as
\[
u^\bmu (s,a) = u(s,a) - \sum_{n\in\mathcal{N}}\mu_n u_n(s,a).
\]
We define a Lagrangian function as
\[
L^{\pi,\bmu}(s)~
	= U^\pi (s) + \sum_{n\in\mathcal{N}} \mu_n (\delta_n- U_n^\pi(s)) \nonumber \\
	= V^{\pi,\bmu}_{\mathcal{M}}(s) + \bmu^\top \bdelta,
\]
where $V^{\pi,\bmu}_{\mathcal{M}}=\mathbb{E} \left[ \sum_{t=0}^\infty (\gamma)^t u^\bmu (s,a) \,\big|\, s^0=s \right]$.
We can reformulate the CMDP to the UMDP as
\begin{equation}
\label{eqn:umdp}
\max_{\pi} \, \min_{\bmu\succeq\mathbf{0}} \, L^{\pi,\bmu}(s)=U^\pi(s)+\sum_{n\in\mathcal{N}}\mu_n(\delta_n-U_n^\pi(s)), \nonumber
\end{equation}
where $\bmu=(\mu_n)_{\forall n\in\mathcal{N}}$, $\succeq$ denotes the elementwise inequality operator, and $\mathbf{0}$ is the zero vector.
This UMDP is a dual problem of the CMDP in \eqref{eqn:cmdp}.
The following theorem from \citep{Altman1998Constrained} implies that the optimal policy to the CMDP can be obtained by solving the UMDP.
\begin{theorem}\label{thm:CMDP=UMDP}
The optimal value of the CMDP in \eqref{eqn:cmdp}, $J^*(s)$, can be obtained by solving the UMDP in \eqref{eqn:umdp} as
\[
L^{\pi^*,\bmu^*}(s)=\max_{\pi} \, \min_{\bmu\succeq\mathbf{0}} \, L^{\pi,\bmu}(s)= \min_{\bmu\succeq\mathbf{0}} \, \max_{\pi} \, L^{\pi,\bmu}_{\mathcal{M}}(s).
\]
The policy $\pi^*$ of the UMDP is optimal for the CMDP if and only if $L^{\pi^*,\bmu^*}(s)= \min_{\bmu\succeq\mathbf{0}} \, L^{\pi^*,\bmu}(s)$.
\end{theorem}
This theorem implies that we can solve the CMDP by finding the optimal policy and the optimal Lagrangian multiplier together.

First, to find the optimal Lagrangian multipliers, we can update the multipliers by using the stochastic subgradient algorithm \citep{ghadimi2012optimal} as
\begin{equation}\label{eqn:multiplier_update}
\mu_n^{t+1}=\left[\mu_n^t-\alpha^t\left(\delta_n-u_n^t\right)\right]^+, ~ \forall n\in\mathcal{N},
\end{equation}
where $\alpha^t$ is the step size in timestep $t$.
Also, we need to find the optimal policy for a given $\bmu$, which can be defined by using the optimal action-value function $Q^{*,\bmu}$ as
\[
\pi^{*,\bmu}(s)=\argmax_{a\in\mathcal{A}(s)}Q^{*,\bmu}(s,a),~\forall s\in\mathcal{S}.
\]
We can find the optimal policy by finding the optimal action-value function defined by using the optimal state-value function $L^{*,\bmu}$ as
\[
Q^{*,\bmu}(s,a)=u^\bmu(s,a)+\gamma\sum_{s'\in\mathcal{S}} \mathcal{P}_{s,a,s'}L^{*,\bmu}(s').
\]
By updating the multipliers, $\bmu$, as in \eqref{eqn:multiplier_update} and learning the optimal action-value function, $Q^{*,\bmu}$, we can solve the CMDP in \eqref{eqn:cmdp}.

\section{DESCRIPTION OF SIMULATION ENVIRONMENT}
We here describe the simulation environments in different scenarios.
We developed a PyTorch-based simulator in python for each scenario and run it on a computer with Intel i7-8700 3.2GHz CPU and 24 GB RAM without any GPU.

\subsection{Description of Simple Explanatory Scenario}
In this simple explanatory scenario, a dynamic scheduling system with $N$ items is considered.
We denote the price of item $n$ and the quantity of item $n$ to sell in timestep $t$ by $p_n^t\in[0,1]$ and $g_n^t\in\{0,1,2,3,4\}$, respectively.
In each timestep, $p_n$ and $g_n$ randomly vary in an i.i.d. manner and the system should select one of the items to sell it.
In specific, $p_n$ is generated according to a uniform distribution and $g_n$ is generated from the set of candidate values with equally distributed probabilities.
We assume that if an item is selected, it will be sold-out during the scheduled timestep.
Then, the reward of timestep $t$ is derived as $r^t=p_{n^t}^tg_{n^t}^t$, where $n^t$ is the selected item in timestep $t$.
The goal of the scheduler is to maximize the reward.
In this scenario, the price and quantity of item to sell are the feature information used for the state.
In the conventional policy structure, the state is defined as $s^t=(p_1,g_1,...,p_N,q_N)$ and the action is defined as $a^t=n^t$.
For the descriptive policy structure, we uniformly partition the space of the price into 4 subsets (i.e., $\{[0,0.25),[0.25,0.5),[0.5,0.75),[0.75,1]\}$).
Then, the descriptive state is defined as $4\times 5$ matrix whose element $h=(h_p,h_g)$ is determined as in (5) of the paper, where $h_p$ and $h_g$ is the indices of the price and the quantity, respectively.

In the simulation, we consider three different system characteristics A, B, and C that have the different number of items, $N$, as $\{2,6,10\}$, respectively.
For evaluation, we consider a conventional policy (Conv-P) for each system characteristics that is tailored to the system characteristics, a descriptive policy (Desc-P), and the optimal policy (Opt-P) for each system characteristics.
We consecutively perform the simulation for system characteristics A, B and C in an online manner with $10^5$ time-slots for each system characteristics,
but the descriptive policy is not trained during system characteristics C to clearly show its system-agnostic capability.
On the other hand, three Conv-Ps are trained during their corresponding system characteristics.

We set the discount factor $\gamma$ to be 0.9.
For the DQNs of the conventional policy and the descriptive policy, we use a feed-forward network composed of 2 hidden layers with 100 units.
The deep RL method in \citep{mnih2015human} is applied to learn the DQNs.
Specifically, we use the $\epsilon$-greedy exploration method, where $\epsilon$ is defined as
\begin{equation}
\epsilon=\left\lbrace \begin{array}{l}
t^{-1},~t \geq 10^4 \\
0.1,~\textrm{otherwise}
\end{array}
\right..
\end{equation}
Also, we set the experience buffer size for the experience replay to be 300 and the target update interval for the fixed target-Q to be 100.
The batch size for training is set to be 30 and the training interval is set to be 10.
We use the Adam optimizer with $10^{-3}$ learning rate.

\subsection{Description of Realistic Application Scenario in Wireless Networks}
\begin{paragraph}{Problem Description}


In this realistic application scenario, we consider a user scheduling and power allocation problem in a typical wireless network consisting of one base station
The network consists of one base station (BS) and $N$ users.
The BS schedules a user and chooses a transmission power to serve the user over discrete time horizon and in each timestep (typically, time-slot in wireless networks) the wireless channel condition of each user randomly varies as a random disturbance.
We assume that the wireless channels between the BS and users are unchanged during a time-slot and follow the Markov property.
The goal of the scheduling is to minimize the average transmission power while satisfying the minimum average data rate requirements.

We denote the scheduled user in timestep $t$ by $n^t$ and the chosen transmission power by $p^t\in\mathcal{P}=\{p_1,p_2,...,p_{max}\}$, where $\mathcal{P}$ is the set of the candidate transmission power levels.
The wireless channel gain of user $n$ in time-slot $t$ is denoted by $\xi_n^t$.
Then, the instantaneous achievable data rate of user $n$ in time-slot $t$ is determined by the Shannon capacity formula as
\[
r_n^t=\left\lbrace \begin{array}{l}
W\log_2{\left(1+\frac{\xi_n^tp^t}{W N_0}\right)},~\textrm{if user $n$ is scheduled in time-slot $t$} \\
0,~\textrm{otherwise}
\end{array}
\right..
\]
where $W$ is the bandwidth of the system and $N_0$ is the noise power density.
Then, the minimum average data rate of user $n$ is defined as
\begin{equation}
\label{eqn:const:average_datarate}
\lim_{t\rightarrow\infty}{\frac{1}{t}\sum_{\tau=0}^{t-1}{r_n^\tau}} \geq \delta_n,~\forall n\in\mathcal{N}.
\end{equation}
In this system model, the wireless channel gain of each user is the feature information used as the state and the scheduled user index and the transmission power are used as the action.
Hence, the state is defined as $s^t=(\xi_1,...,\xi_N)$ and the action is defined as $a^t=(n^t,p^t)$.

We now formulate a user scheduling and power allocation problem in a form of the CMDP as follows:
\[
\minimize_{\pi:\mathcal{S}\rightarrow\mathcal{A}} ~\mathbb{E}\left[\sum_{t=0}^{\infty}{(\gamma)^tp(s^t,\pi(s^t))} \right]
\subjecto ~\mathbb{E}\left[\sum_{t=0}^{\infty}{(\gamma)^tr_n(s^t,\pi(s^t))} \right]\geq \bar{\delta}_n,~\forall n\in\mathcal{N},
\]
where the constraints are derived from the average data rate requirement in \eqref{eqn:const:average_datarate} considering the discount factor.
Since the constraint in \eqref{eqn:const:average_datarate} does not consider the discount factor $\gamma$, we need to transform it into the discounted data rate requirement as in \citep{Mastronarde2013Joint}.
The discounted constraint constant $\bar{\delta}_n$ is calculated as $\delta_n/(1-\gamma)$.
This CMDP follows the dynamic scheduling problem with constraints.
Hence, to solve this problem, we define the Lagrangian multiplier of the data rate requirement of user $n$ as $\mu_n$ and update it by using the stochastic subgradient algorithm as
\[
\mu_n^{t+1}=\left[\mu_n^t-\alpha^t(r_n^t-\bar{\delta}_n)\right]^+.
\]
\end{paragraph}

\begin{paragraph}{Simulation Scenario Description}
In the simulation, we set the bandwidth of the system, $W$, to be 5 MHz and the noise spectral density, $N_0$, to be -106 dBm/Hz.
The set of candidate transmission power level is calculated by uniformly discretizing $[0,10]$ W into 5 values.
For wireless channels, we set the pathloss exponent to be 3.76 and consider a log-normal shadowing with 10 dB standard deviation.
In the conventional policy structure, the augmented state defined as $\underline{s}^t=(\xi_1,\mu_1,...,\xi_N,\mu_N)$ and the action defined above are used.
For the descriptive policy structure, we define the space of the wireless channel gain of each user by $[-30,-50]$ dB and project the wireless channel gain to the space.
We then uniformly partition the space into 5 subsets.
Similarly, we define the space of the Lagrangian multiplier of each user by $[0,2]$ and project the multiplier to the space.
We partition the space into 10 subsets.
When partitioning the space of the multiplier, we make the partition more fine in a small value region since the values of the multipliers are typically distributed in the region.
Here is an example: $\{[0,0.025),[0.025, 0.099),[0.099,0.22),,...,[1.58,2]\}$.
Then, the descriptive state is defined as $5\times 10$ matrix whose element $h=(h_\xi,h_\mu)$ is determined as in (5) of the paper, where $h_\xi$ and $h_\mu$ is the indices of the wireless channel gain and the multiplier, respectively.

\begin{table}[!t]
\fontsize{9pt}{9pt}\selectfont
\centering
\caption{System characteristics for realistic application in wireless networks}
\label{table:topologies}
\begin{tabular}{c|c|c||c}
\hline
& \multicolumn{2}{c||}{\textbf{Online learning}} & \textbf{Pre-trained} \\
\hline
& \textbf{System A} & \textbf{System B} & \textbf{System C} \\
\hline \hline
No. of users & 4 & 9 & 20 \\
\hline
\makecell{Distance from BS} & \makecell{20 m (1 users)\\50 m (2 users)\\80 m (1 users)} & \makecell{20 m (3 users)\\50 m (3 users)\\80 m (3 users)} & \makecell{20 m (5 users)\\50 m (10 users)\\80 m (5 users)} \\
\hline
Data rate reqs. & 1 Mbps & 0.5 Mbps & 0.2 Mbps \\
\hline
\end{tabular}
\end{table}

In the simulation, we consider three different system characteristics A, B, and C that have the different number of users, the distance of each user from the BS, and the minimum average data rate requirements. Details of the system characteristics are provided in Table \ref{table:topologies}.
The simulation is conducted during $10^6$ time-slots for each system characteristics.
For system characteristics A and B (an online learning case), we consecutively perform the simulation as in the previous simple explanatory scenario.
For system characteristics C (a pre-trained case), we train a Conv-P in advance until it converges (about $10^7$ time-slots).
Moreover, the Desc-P trained during system characteristics A and B is used as in the previous scenario.

We set the discount factor $\gamma$ to be 0.9.
For the DQNs of the conventional policy and the descriptive policy, we use a feed-forward network composed of 4 hidden layers with 300 units.
We use the deep RL method in \citep{mnih2015human} as well, and the parameters are set to be the same above except for $\epsilon=0.1$ and the learning rate $10^{-4}$.
\end{paragraph}

\subsection{Description of Simulation Environment for Multiple Dynamic Scheduling Systems with Federated Learning}
We now describe the simulation environment for federated learning (FL) in which multiple dynamic scheduling systems exist.
We apply FL to both simple explanatory and realistic application scenarios.
In both scenarios, we consider two identical systems having each system characteristics so that we have total six systems.
We also consider one global server to aggregate the DQN of each system via federated learning.
Then, each system runs in parallel while all systems share the same timestep.
At the end of the timestep in which the local DQNs are trained, the local DQN for the Desc-P of each system is simply aggregated by the global server and distributed to the systems as in \citep{mcmahan2017communication}.
We use the same parameters with the corresponding scenarios.

\section{ADDITIONAL EXPERIMENTAL RESULTS}

\subsection{Simple Explanatory Scenario}

\begin{wrapfigure}{r}{0.35\textwidth}
\vspace{-1em}
\centering
\includegraphics[width=0.3\textwidth]{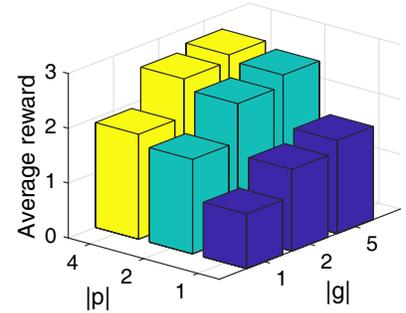}
\caption{Average reward varying fineness of descriptive states.}

\label{fig:sc1_generalization}
\end{wrapfigure}

\textbf{Impact of Fineness of Descriptive State (Impact of Generalization)}~~~
We provide the average rewards of system characteristics C achieved by the Desc-Ps with different fineness of descriptive state in Fig. \ref{fig:sc1_generalization}.
In the figure, the label $|p|$ denotes the number of subintervals for $p_n\in[0,1]$ and the label $|g|$ denotes the number of subsets for $g_n\in\{0,1,2,3,4\}$.
To adjust the fineness, we vary $|p|$ as 1,2,4 and $|g|$ as 1,2,5.
For $p_n$, we uniformly partition the interval $[0,1]$ according to $|p|$.
Also, we use the subsets $\{0,1,2\}$ and $\{3,4\}$ for $|g|=2$.
From the figure, we can see that the descriptive state becomes finer, the average reward increases.
This is natural since the finer descriptive state implies less generalization of the states and actions. (This is implicitly shown in Theorem 1 as well.)
Also, we can see that the increment of the average reward decreases as the descriptive state becomes finer and finer.
Since finer descriptive states require more complexity of learning, this clearly shows a need of controlling the trade-off between the performance and complexity from the generalization.

\begin{figure}[!h]
\centering
\includegraphics[width=.78\textwidth]{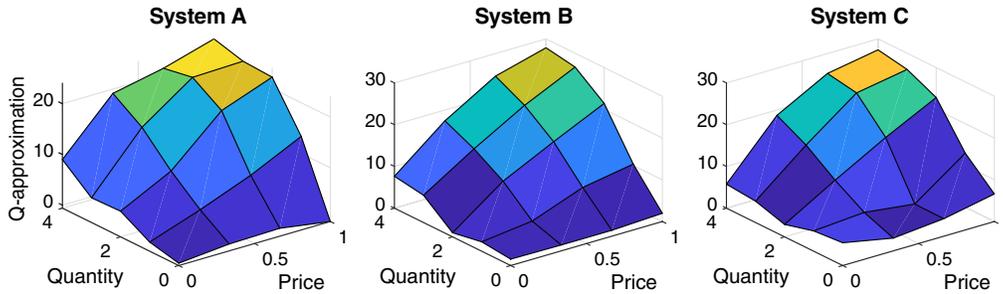}
\caption{Q-approximations of the descriptive policies learned from different systems.}
\label{fig:sc1_q_approx}
\end{figure}

\begin{paragraph}{Q-Approximations of Descriptive Policy}
We train three Desc-Ps by using the experiences from system characteristics A, B, and C, respectively, to clearly show the capability of the Desc-P to learn a system-agnostic scheduling principle.
In Fig. \ref{fig:sc1_q_approx}, the Q-approximation of the Desc-P for each system is provided.
From the figure, we can see that the Q-approximations have an identical trend that is aligned with the optimal scheduling principle regardless of system characteristics despite the Desc-Ps are trained from different system characteristics.
This clearly shows the system-agnostic capability of the Desc-P by learning the system-agnostic scheduling principle, which explains how the Desc-P learned from one system can be adopted to the other system.
\end{paragraph}

\begin{figure}[!h]
\centering
\includegraphics[width=0.6\textwidth]{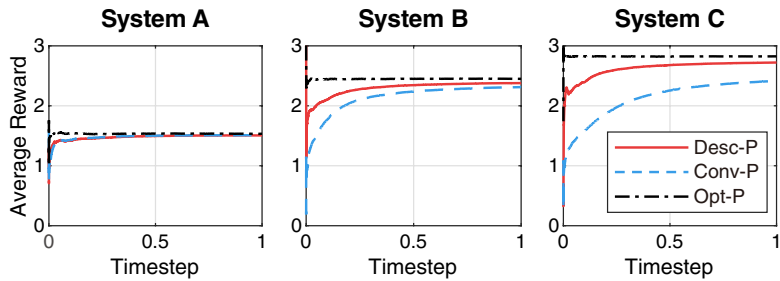}
\caption{Average reward in each system characteristics.}
\label{fig:sc1_learningspeed}
\end{figure}

\textbf{Impact of System Complexity}~~
Here, we train Desc-P and Conv-P for each system characteristics.
In this simulation, to show the learning speed of the policies, when learning them in one system characteristics, the experiences from the system are used only.
Since the system characteristics have different number of items, they have different system complexity.
In the Conv-P, as the number of items increases, the policy structure becomes more complex since the state and action spaces are enlarged according to the number of items (see equation (1) and (2) in the main paper).
On the other hand, the Desc-P has an identical policy structure regardless of the number of items.
Hence, the learning complexity of the Conv-P depends on the system complexity more directly compared with that of the Desc-P.
This is clearly shown in Fig. \ref{fig:sc1_learningspeed}.
The figure provides the average rewards of the Desc-P, Conv-P, and Opt-P in each system characteristics.
We can see that in system characteristics A with 2 items, both Desc-P and Conv-P have the similarly fast learning speed due to the low system complexity.
However, as the number of items increases in system characteristics B (6 items) and C (10 items), the learning speed of the Conv-P becomes significantly slower.
On the other hand, the Desc-P has the similar learning speed in both system characteristics B and C.
This clearly shows that the advantage of the descriptive policy structure in terms of learning speed.


\begin{figure}[H]
\begin{minipage}{0.5\textwidth}
\centering
\includegraphics[width=0.45\textwidth]{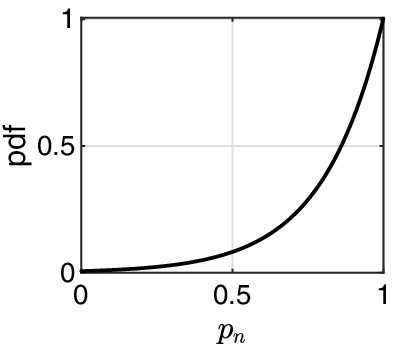}
\caption{Probability density of $p_n$.}
\label{fig:sc1_partitioning}
\end{minipage}
\hfil
\begin{minipage}{0.5\textwidth}
\centering
\includegraphics[width=0.45\textwidth]{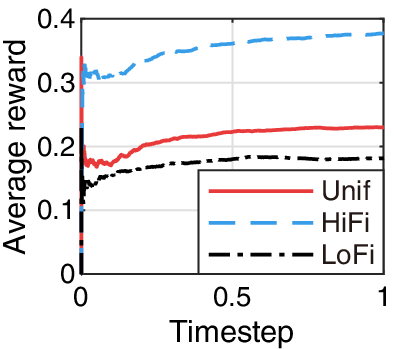}
\caption{Average reward.}
\label{fig:sc1_partitioning_reward}
\end{minipage}
\end{figure}

\textbf{Impact of Partitioning Strategy}~~
The partitioning strategy for descriptive state affects the performance of Desc-P.
To clearly show this, we use a truncated exponential distribution to generate $p$.
In specific, the price of item $n$, $p_n$, is generated as $1-\phi/5$, where $\phi$ is the truncated exponential random variable whose probability density is given as $e^{-x}/(1-e^{-5})$.
Then, the probability distribution of $p_n$ is provided in Fig. \ref{fig:sc1_partitioning}.
Also, to emphasize the higher price, we use the reward given by $r=p_n^{100}g_n$, where $n$ is the selected item.
In this scenario, discriminating the higher price is important to achieve the larger average reward.
We consider three partitioning strategies that partition $[0,1]$ into 4 subintervals.
In unifrom partitioning (Unif), the interval the interval is uniformly partitioned.
In finer partitioning in higher-region (HiFi), the interval is partitioned as $\{[0,0.5),[0.5,0.75),[0.75,0.9),[0.9,1]\}$, and in finer partitioning in low-region (LoFi),  the interval is partitioned as $\{[0,0.1),[0.1,0.25),[0.25,0.5),[0.5,1]\}$.
Fig. \ref{fig:sc1_partitioning_reward} provides the average rewards of Desc-P with the different partitioning strategies in system characteristics C.
In the figure, HiFi achieves the best performance by discriminating the higher prices.
This clearly shows that the partitioning strategies affect the performance of the Desc-P.

\subsection{Realistic Application Scenario in Wireless Networks}

\begin{figure}[!h]
\centering
\includegraphics[width=1\textwidth]{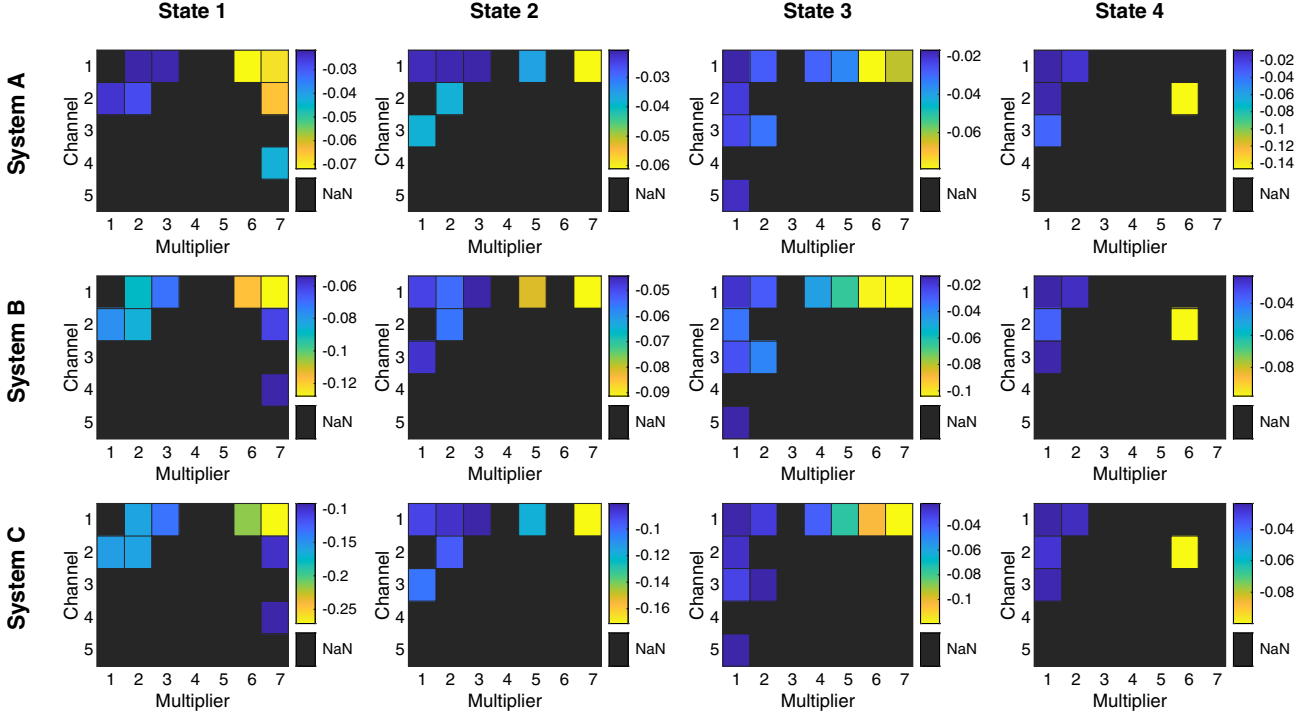}
\caption{Q-approximations of the different descriptive states with the descriptive policies learned from each system.}
\label{fig:sc2_q_approx}
\end{figure}

\begin{paragraph}{Q-Approximations of Descriptive Policy}
The descriptive policy for the realistic application scenario in wireless networks can learn the scheduling principle for user scheduling.
To show this, we can compare the Q-approximations of given descriptive states as in the results with the simple explanatory scenario.
The realized descriptive state in this realistic scenario is sparse and the descriptive state whose all elements have a value of one is impractical since the descriptive state space is larger than that in the simple scenario.
(The number of users in the network is smaller than the number of the conditions in the descriptive state.)
Hence, to show the Q-approximations, here we use the descriptive states that are generated in the simulation instead of using the descriptive states whose all elements has a value of one as in Fig. \ref{fig:sc1_q_approx}.

Similarly to the simple scenario, we train three Desc-Ps by using the experiences from system characteristics A, B, and C, respectively.
We then calculate the Q-approximations by using four different descriptive states.
In the descriptive policy, the conditions in which no user belongs cannot be chosen as the descriptive action.
We illustrate such conditions as \texttt{NaN}.
Also, for each condition, the different Q-approximations can be calculated for different transmission power.
Hence, to merge them, we use the minimum Q-approximation value among them since in DRL, the action having the minimum Q-approximation value is chosen among all the candidate actions.

We provide the Q-approximations in Fig. \ref{fig:sc2_q_approx}.
In the figure, the lower channel subinterval index implies the better channel condition, and the lower multiplier subinterval index implies the lower multiplier.
The multiplier subintervals have the indices from 1 to 10, but the subintervals from 8 to 10 have too large values so as that the multiplier that belongs to such subintervals hardly appears.
Hence, in the figure, we illustrate the multiplier subinterval indices only from 1 to 7.
From the figure, we can see that the Q-approximations for each descriptive state have an identical trend regardless of system characteristics and descriptive states: a user with the better channel condition and the higher multiplier should be scheduled.
This trend is completely aligned to the domain knowledge of wireless communications on user scheduling.
Hence, these results clearly shows the system-agnostic capability of the Desc-P as in the simple explanatory scenario.
\end{paragraph}

\begin{figure}[!h]
\centering
\includegraphics[width=0.55\textwidth]{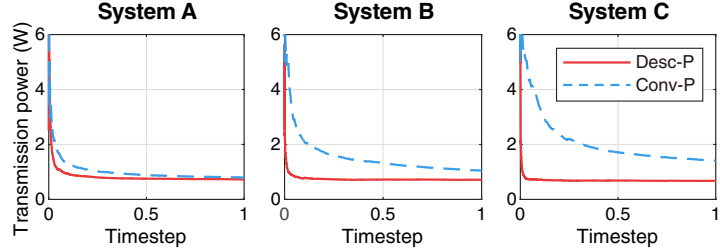}
\caption{Average transmission power in each system characteristics.}
\label{fig:sc2_learningspeed}
\end{figure}

\begin{paragraph}{Impact of System Complexity}
Here, we train Desc-P and Conv-P for each system characteristics.
As in the simulation with the simple explanatory scenario, when learning them in one system, we use the the experiences from the system only to show the learning speed of the policies.
Each system characteristics has different number of users as described in Table \ref{table:topologies}, which incurs different system complexities.
The Conv-P becomes more complex as the number of items increases, while the Desc-P does not.
From Fig. \ref{fig:sc2_learningspeed}, this difference between the Conv-P and the Desc-P is clearly illustrated.
As the number of users increases, the learning speed of the Conv-P becomes significantly slower.
On the other hand, that of the Desc-P is similar, and even becomes slightly faster.
This is because the descriptive state from the system with small number of users is too sparse to effectively learn the Desc-P.
For example, in system characteristics A, at most four elements can have a value of one in the descriptive state.
These results clearly show the advantage of the descriptive policy structure on learning speed compared with the conventional policy structure.
\end{paragraph}

\subsection{Multiple Dynamic Scheduling Systems Scenario with Federated Learning}
The effectiveness of federated learning (FL) can be clearly shown in a system that is complex enough to be learned by multiple learners.
Hence, for this scenario with FL, we consider the descriptive policy for the realistic wireless networks in practical use.
For this, we use a descriptive policy structure that has the descriptive states and actions with more finer partitions than the previous scenario as follows: the wireless channel gain space is partitioned into 15 subsets, the space of the Lagrangian multiplier is partitioned into 12 subsets, and the number of the transmission power levels is given by 20.
We consider a system composed of one central unit and three independent subsystems with system characteristics A, B, and C used in the previous simulations.
Then, the descriptive policy is applied to each subsystem for dynamic scheduling.
Also, FL is adopted to this system as described in Section 3.2 of the main paper, where each subsystem plays a role of the local learner and the central unit aggregates the DQN of each local learner.

\begin{figure}[!h]
\centering
\includegraphics[width=0.95\textwidth]{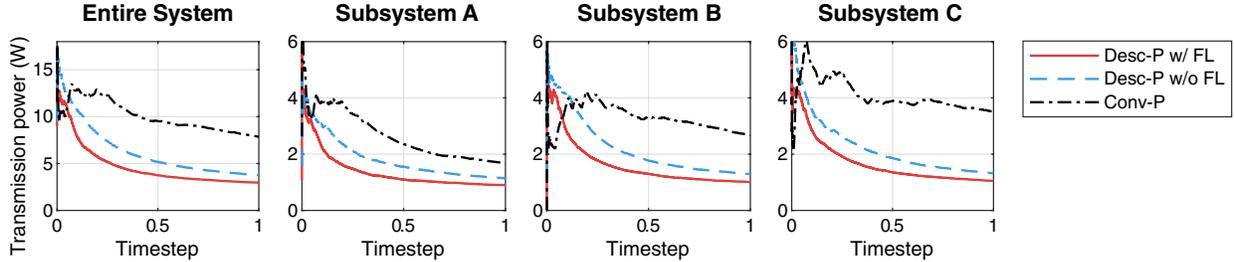}
\caption{Average transmission power in each system characteristics and the entire system.}
\label{fig:fl_speed}
\end{figure}

Fig. \ref{fig:fl_speed} provides the average transmission power in each subsystem and the entire system with different policies.
To show the effectiveness of FL, we provide the average transmission power with the Desc-P with FL and the Desc-P without FL.
We also provide that with the Conv-P to which FL cannot be applied.
From the figure, we can see that the Desc-P achieves the faster learning speed than the Conv-P as shown in the previous results.
Note that FL cannot improve the converged Desc-P since the Desc-P learns the scheduling principle that is system-agnostic.
However, as shown in the figure, FL accelerates the learning speed of the Desc-P by aggregating the experiences from the subsystems.
This clearly shows the advantage of FL over multiple dynamic scheduling systems on learning speed.

\end{document}